\documentclass{article}

\usepackage[preprint, nonatbib]{nips_2018}
\usepackage{booktabs} 
\usepackage[justification=centering]{caption}
\usepackage{amsmath}
\usepackage{lipsum}
\usepackage{balance}
\usepackage{graphicx}
\usepackage{hyperref}

\renewcommand{\vec}[1]{\mathbf{#1}}

\begin{document}
\title{Evolving Differentiable Gene Regulatory Networks}

\author{DG Wilson\\
University of Toulouse, IRIT - CNRS - UMR5505, Toulouse, France\\
\texttt{dennis.wilson@irit.fr}
\AND Kyle Harrington\\
Virtual Technology + Design, University of Idaho, Moscow, ID 83844, USA\\
\texttt{kharrington@uidaho.edu}
\AND Sylvain Cussat-Blanc\\
University of Toulouse, IRIT - CNRS - UMR5505, Toulouse, France\\
\texttt{sylvain.cussat-blanc@irit.fr}
\AND Herv\'e Luga\\
University of Toulouse, IRIT - CNRS - UMR5505, Toulouse, France\\
\texttt{herve.luga@irit.fr}
}



\maketitle

\begin{abstract}
  Over the past twenty years, artificial Gene Regulatory Networks (GRNs) have
shown their capacity to solve real-world problems in various domains such as
agent control, signal processing and artificial life experiments. They have also
benefited from new evolutionary approaches and improvements to dynamic which
have increased their optimization efficiency. In this paper, we present an
additional step toward their usability in machine learning applications. We
detail an GPU-based implementation of differentiable GRNs, allowing for local
optimization of GRN architectures with stochastic gradient descent (SGD). Using
a standard machine learning dataset, we evaluate the ways in which evolution and
SGD can be combined to further GRN optimization. We compare these approaches
with neural network models trained by SGD and with support vector machines.

\end{abstract}

\section{Introduction}
\label{sec:intro}

Artificial Gene Regulatory Networks (GRNs) have varied in implementation since
their conception, with initial boolean network representations that directly
encode connections giving way to proteins that interpolate their connection
using exponential or other functions. Following inspiration from biology, these
models have been optimized using genetic algorithms, with advances in GRN
representation often involving improvements to the evolvability of the
representation \cite{kuo_evolving_2004}, \cite{Cussat-Blanc2015a},
\cite{disset2017comparison}. 

Artificial GRNs were first proposed using a binary encoding of proteins with
specific start and stop codons, similar to biological genetic encoding
\cite{Banzhaf2003}. GRNs have since been used in a number of domains, including
robot control \cite{Joachimczak2010a}, signal processing \cite{Joachimczak2010},
wind farm design, \cite{Wilson2013}, and reinforcement learning
\cite{Cussat-Blanc2015}. Finding similar use to their biological inspiration,
GRNs have controlled the design and development of multi-cellular creatures
\cite{Cussat-Blanc2008} and of artificial neural networks (ANNs)
\cite{Wrobel2012}. Evolution and dynamics were recently improved in
\cite{Cussat-Blanc2015a,disset2017comparison}. Artificial GRNs have also been
used to investigate a number of questions in the context of evolution. A strong
relationship was shown between their robustness against noise and against
genetic material deletions \cite{rohlf_emergent_2009}. Redundancy of genetic
material was proven to enhance evolvability up to a point
\cite{schramm2012redundancy} and it was shown that modular genotypes can emerge
when GRNs are subjected to dynamic fitness landscapes \cite{lipson_origin_2002}.

In this paper, we present a model that takes the process of optimizing GRNs one
step further, by introducing the differentiable Gene Regulatory Network, a GRN
capable of learning. We show that differentiable GRNs benefit from a combination
of evolution and learning. The availability of modern platforms capable of
performing gradient descent with complex functions has enabled this work. A
differentiable GRN written using
TensorFlow\footnote{https://www.tensorflow.org/} is presented. For ease of
comparison with and incorporation into existing learning models, especially deep
neural networks, the GRN has been implemented as a Keras
layer.\footnote{https://keras.io/} This implementation is available in an
open-source repository.\footnote{https://github.com/d9w/pyGRN}

This paper also aims at studying the relationship between evolution and learning
using GRNs. While this constitutes a first look with GRNs (to the best of our
knownledge), ANNs have been used to understand both the Baldwin effect and
Lamarckian evolution. ANNs have been evolved with the option of allowing certain
weights to be learned, demonstrating that learning can improve evolution's
ability to reach difficult parts of the search space \cite{hinton1987learning}.
Lamarckian evolution has been used to combine evolution of the structure of a
Complex Pattern Producing Network and learning of the network's weights
\cite{fernando2016convolution}. This work intends to explore the relationship
between evolution and learning in the context of GRNs, which are detailed next.

Plasticity is an adaptive response variation that allows an organism to respond
to environmental change. The evolutionary advantage of plasticity was first
proposed in \cite{baldwin1896new} and is now referred to as the Baldwin Effect.
The Baldwin Effect suggests that learned/adaptive behaviors which have fitness
advantages can facilitate the genetic assimilation of equivalent behaviors by
subsequent generations. The Baldwin Effect has been studied in a number of
systems, including RNA \cite{ancel2000plasticity}, neural networks
\cite{schemmel2006implementing}, and theoretical models
\cite{ancel2000undermining}. We suggest that GRNs which can change or ``learn''
over their lifetime may be influenced by the Baldwin effect; evolutionary
selection based upon learning may produce a population in parts of the
evolutionary search space which are difficult to reach without learning.

To this end, we first present the differentiable GRN model, transforming standard
GRN equations into a series of differentiable matrix operations, and use this
model to address some questions concerning the relationship between evolution
and learning. The GRN model is uniquely poised to study aspects of this complex
relationship, as evolutionary methods for the model have been well studied and
as the entire genome is differentiable.


\section{Implementation}
\label{sec:grn}


A GRN is composed of multiple artificial proteins, which interact via evolved
properties. These properties, called tags, are: the protein \emph{identifier},
encoded as a floating point value between 0 and 1; the \emph{enhancer
  identifier}, encoded as a floating point value between 0 and 1, which is used
to calculate the enhancing matching factor between two proteins; the
\emph{inhibitor identifier}, encoded as a floating point value between 0 and 1,
which is used to calculate the inhibiting matching factor between two proteins
and; the \emph{type}, either \emph{input}, \emph{output}, or \emph{regulator},
which is a constant set by the user and is not evolved.

Each protein has a concentration, representing the use of this protein and
providing state to the network. For \emph{input} proteins, the concentration is
given by the environment and is unaffected by other proteins. \emph{output}
protein concentrations are used to determine actions in the environment; these
proteins do not affect others in the network. The bulk of the computation is
performed by \emph{regulatory} proteins, an internal protein whose concentration
is influenced by other \emph{input} and \emph{regulatory} proteins.

We will first present the classic computation of the GRN dynamics, using
equations found to be optimal in \cite{disset2017comparison} on a number of
problems. Following this overview, we will present the conversion of these
equations into a set of differentiable matrix operations.

The dynamics of the GRN are calculated as follows. First, the absolute affinity of a
protein $a$ with another protein $b$ is given by the enhancing factor
$u^{+}_{ab}$ and the inhibiting $u^{-}_{ab}$:
\begin{equation}
\label{eq:affinity_first}
u^{+}_{ij}=u_{size}-|enh_j-id_i|~~;~~u^{-}_{ij}=u_{size}-|inh_j-id_i|
\end{equation}

where $id_x$ is the identifier, $enh_x$ is the enhancer identifier and $inh_x$
is the inhibitor identifier of protein $x$. The maximum enhancing and inhibiting
affinities between all protein pairs are determined and are used to calculate
the relative affinity, which is here simply called the affinity:
\begin{equation}
\label{eq:affinity_main}
A^{+}_{ij}=-\beta u^{+}_{ij}~~;~~A^{-}_{ij}=-\beta u^{-}_{ij}
\end{equation}
$\beta$ is one of two control parameters used in a GRN, both of which are
described below.

These affinities are used to then calculate the enhancing and inhibiting
influence of each protein, following
\begin{equation}
\label{eq:dynamics_1}
g_i=\frac{1}{N}\sum_j^N{c_je^{A^{+}_{ij}}}~~;~~h_i=\frac{1}{N}\sum_j^N{c_je^{A^{-}_{ij}}}
\end{equation}
where $g_i$ (resp. $h_i$) is the enhancing (resp. inhibiting) value for a
protein $i$, $N$ is the number of proteins in the network, $c_j$ is the
concentration of protein $j$.

The final modification of protein $i$ concentration is given by the following
differential equation:
\begin{equation}
\label{eq:dynamics_2}
\frac{dc_i}{dt}=\frac{\delta(g_i-h_i)}{\Phi}
\end{equation}
where $\Phi$ is a function that normalizes the output and regulatory protein
concentrations to sum to 1.

$\beta$ and $\delta$ are two constants that determine the speed of reaction of
the regulatory network. The higher these values, the more sudden the transitions
in the GRN. The lower they are, the smoother the transitions. In this work, the
$\beta$ and $\delta$ parameters are both evolved as a part of the GRN chromosome
and learned in the optimization of the differentiable GRN.

\subsection{Differentiable GRN}


In this representation of the GRN, protein tags are separated into three vectors
based on their function, $\vec{id}$, identifier, $\vec{enh}$, enhancer, and
$\vec{inh}$, inhibitor. This is their form in evolution and the evolved vectors
serve as initial values for learning. Protein concentrations are also
represented as a vector, following the same indexing as the protein tags.
Initial value for protein concentrations is 0.





The protein tags are then tiled to create three matrices. The enhancing and
inhibiting matrices are transposed, and the affinities are then calculated for
each element of the matrix using broadcasting, to be expressed in the form of
the net influence of a protein on another protein, referred to as the protein
signatures, $S$:

\begin{gather}
A^{+} = -\beta |ENH - ID|~~;~~A^{-} = -\beta |INH - ID|~~;~~\\
S = e^{A^{+}}-e^{A^{-}}
\end{gather}

The protein tags $\vec{id}, \vec{enh}, \vec{inh}$ are all optimized by learning,
as well as the $\beta$ and $\delta$ GRN parameters. The protein tags are
constrained during optimization to be in $[0.0, 1.0]$, and $\beta$ and $\delta$
are constrained between the parameter values $[\beta_{min}, \beta_{max}]$ and
$[\delta_{min}, \delta_{max}]$, which, for this run, were both $[0.05, 2.0]$.
Learning therefore directly augments evolutionary search in this work; evolution
also optimizes the protein tags, $\beta$, and $\delta$, and is bound by the same
constraints.

\subsection{GRN layer}

\begin{figure}[t]
  \centering
  \includegraphics[width=0.7\textwidth]{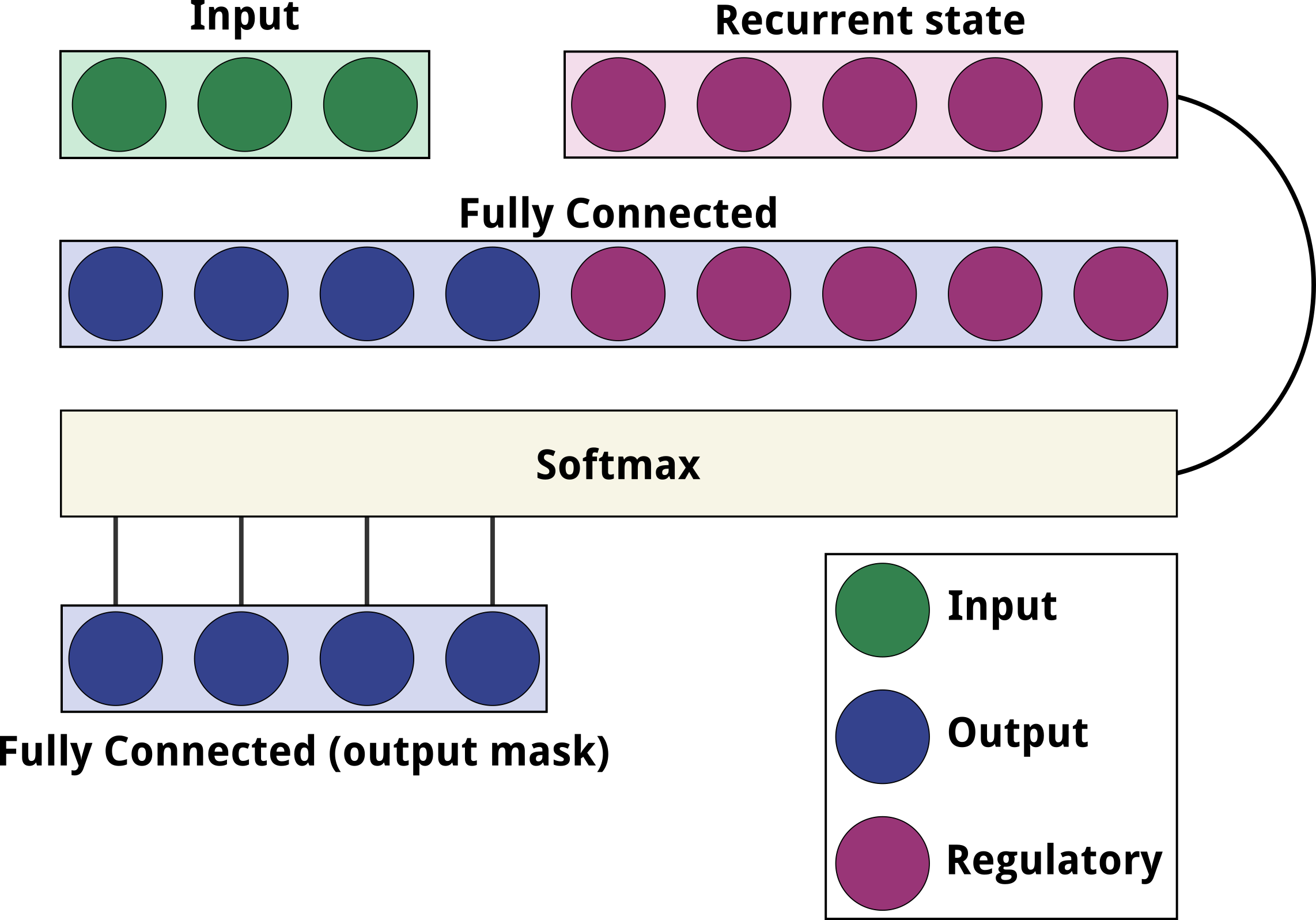}
  \caption{The GRN layer modeled as a composition of other layers}
  \label{fig:grn_layer}
\end{figure}

The differentiable GRN model is encoded as a Keras layer to facilitate use.
Keras is a popular deep learning framework which includes many implemented
layers, such as dense, convolutional, and recurrent neural network layers.
Layers are added to a model class, which provides the interface for learning and
evaluation. Here we consider the GRN layer calculation as compared to common
neural network layers, shown in \autoref{fig:grn_layer}. This is not how
the layer is implemented, but is shown rather to demonstrate the difference
between the GRN and existing layer types.

Input proteins influence output and regulatory proteins according to their
respective signatures, which can be considered as weights; the first part of a
GRN layer is therefore like a fully connected layer containing the output
regulatory proteins. These protein concentrations are also influenced by the
previous concentrations of the regulatory proteins, where the signatures from
the regulatory proteins act as weights to all non-input proteins. Output
proteins, however, do not influence regulatory proteins. The layer is therefore
similar to a classic recurrent layer, in that nodes have non-zero weights
between each other depending on their previous activation, but the state of this
layer is stored in the regulatory proteins alone, not in the output proteins.

\begin{figure}[ht]
  \centering
  \includegraphics[width=0.3\textwidth]{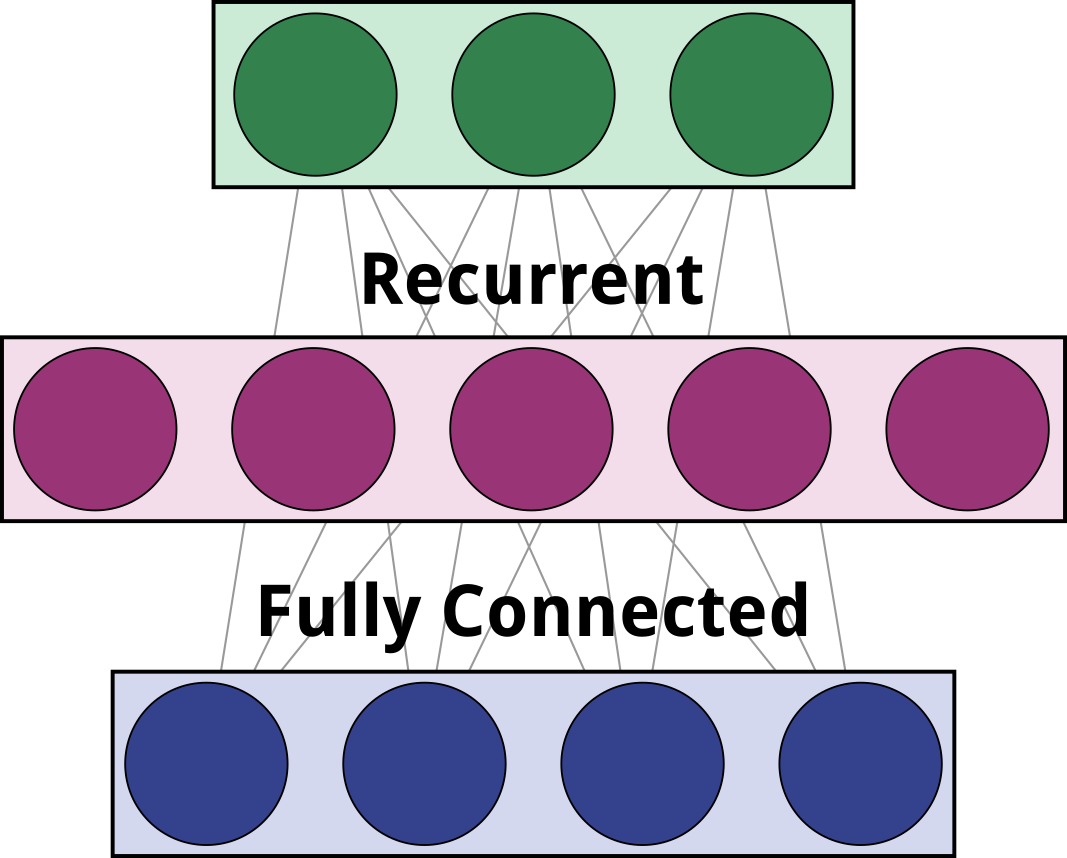}
  \caption{The GRN layer modeled as a composition of other layers, without
    protein concentration normalization}
  \label{fig:grn_layer_no_norm}
\end{figure}

Finally, a large difference between a recurrent layer and the GRN is the
normalization function, which can be modeled as a softmax layer. It is important
to note that the softmax layer is applied before saving the state of the layer;
the regulatory protein concentration which affects the next time-step is already
normalized. Normalization has been shown to be an important part of artificial
GRN evolution and use \cite{disset2017comparison}, however it may confound deep
neural models not accustomed to having a softmax in the interior of the model.
In \autoref{fig:grn_layer_no_norm}, a GRN layer with no normalization is
considered.

Without normalization, the GRN layer resembles a classic recurrent layer
containing the regulatory proteins followed by a fully connected layer
containing the output proteins. However, as input proteins directly influence
output proteins, there must be connections from the inputs to the fully
connected output layer as well.

With these architectural comparisons in mind, the GRN layer resembles common
neural network functions. Proteins behave as rectified linear units, as they are
bound by 0 but otherwise apply no activation transformation. Unlike artificial
neurons, however, GRN proteins have no bias; the activation is simply the sum of
the weighted input. Finally, it is important to note that, while the signature
matrix can be used to represent weights between nodes, the signature matrix was
not optimized during training. Rather, the protein tags $\vec{id}, \vec{enh},
\vec{inh}$ are optimized by learning, as well as the $\beta$ and $\delta$ GRN
parameters. The training is therefore constrained, similar to how convolutional
layers use a kernel to represent multiple weights.

\subsection{Evolution}

The evolutionary method used in this work is the Gene Regulatory Network
Evolution Through Augmenting Topologies (GRNEAT) algorithm
\cite{Cussat-Blanc2015a}. GRNEAT is a specialized Genetic Algorithm for GRNs
which uses an initialization of small networks, speciation to limit competition
to similar individuals, and a specialized crossover which aligns parent genes
based on a protein distance when creating the child genome. It has been shown to
improve results over a standard GA when evolving GRNs on a variety of tasks.

\section{Experiments}
\label{sec:experiments}


The following experiments are intended to demonstrate the capabilities of a
differentiable GRN and to better understand the relationship between evolution
and learning of GRNs. The GRN layer is evaluated on the Boston Housing dataset
\footnote{https://archive.ics.uci.edu/ml/machine-learning-databases/housing/}, a
classic machine learning dataset. data are normalized between $[0.0, 1.0]$ on
each feature. 25\% of the data are reserved for testing, using the same split
for all runs.

\begin{table}[!h]
  \centering
  \begin{tabular}{l l l l}
    \multicolumn{2}{c}{Evolution} & \multicolumn{2}{c}{Learning}\\\hline
    population size & 50 & lr & 0.001\\
    crossover & 0.25 & $\beta_1$ & 0.9\\
    mutation & 0.75 & $\beta_2$ & 0.999\\
    $p_{add}$ & 0.5 & $\epsilon$ & 1e-8\\
    $p_{modify}$ & 0.25 & batch size & 32\\
    $p_{delete}$ & 0.25
  \end{tabular}
  \caption{Evolutionary and learning parameters. GRNEAT parameters were based on
    those used in \cite{disset2017comparison} and learning parameters are the
    defaults in Keras.}
  \label{tab:params}
\end{table}

Mean squared error (MSE) is used as the evolutionary fitness and loss metric for
gradient descent. The evolutionary fitness is the MSE of the GRN after a number
of learning epochs, which indicate one pass through the dataset. Evolution with
three training periods are compared: no learning (0 epochs), minimal learning (1
epoch) and learning (10 epochs). 10 epochs was chosen as the maximum learning
period based on computational cost.

After evolution, best individuals from different generations across the
different evolutions are compared over a longer learning time. This is in order
to understand the possible benefits of evolution when using gradient descent for
optimization. The best GRNs are trained for 200 epochs, at which point there is
clear convergence.


Finally, evolved and trained GRNs are compared to a large random GRN, two neural
network models and to a Support Vector Machine (SVM). The large GRN has 50
nodes, the maximum possible during evolution. The first neural network model
consists of a single fully-connected RNN layer with $N$ nodes, followed by a
densely connected layer with $n_{output}$ nodes. The second neural network model
consists of three densely connected layers, the first with 50 nodes, the second
with 10, and then finally the output layer with $n_{output}$ nodes. The SVM used
a radial basis function kernel.

Training for both the GRN models and compared models uses MSE and the Adam
optimizer with default Keras parameters \cite{kingma2014adam}. Adam was
determined to perform better than baseline stochastic gradient descent when
optimizing the GRN model on the Boston dataset, and is a standard choice for
deep neural network optimization.

The full set of parameters used in evolution and learning are given in
\autoref{tab:params}. The number of generations was chosen based on evolutionary
convergence of the 0 epoch evolution for each dataset.

\begin{figure*}[ht]
  \includegraphics[width=0.5\textwidth]{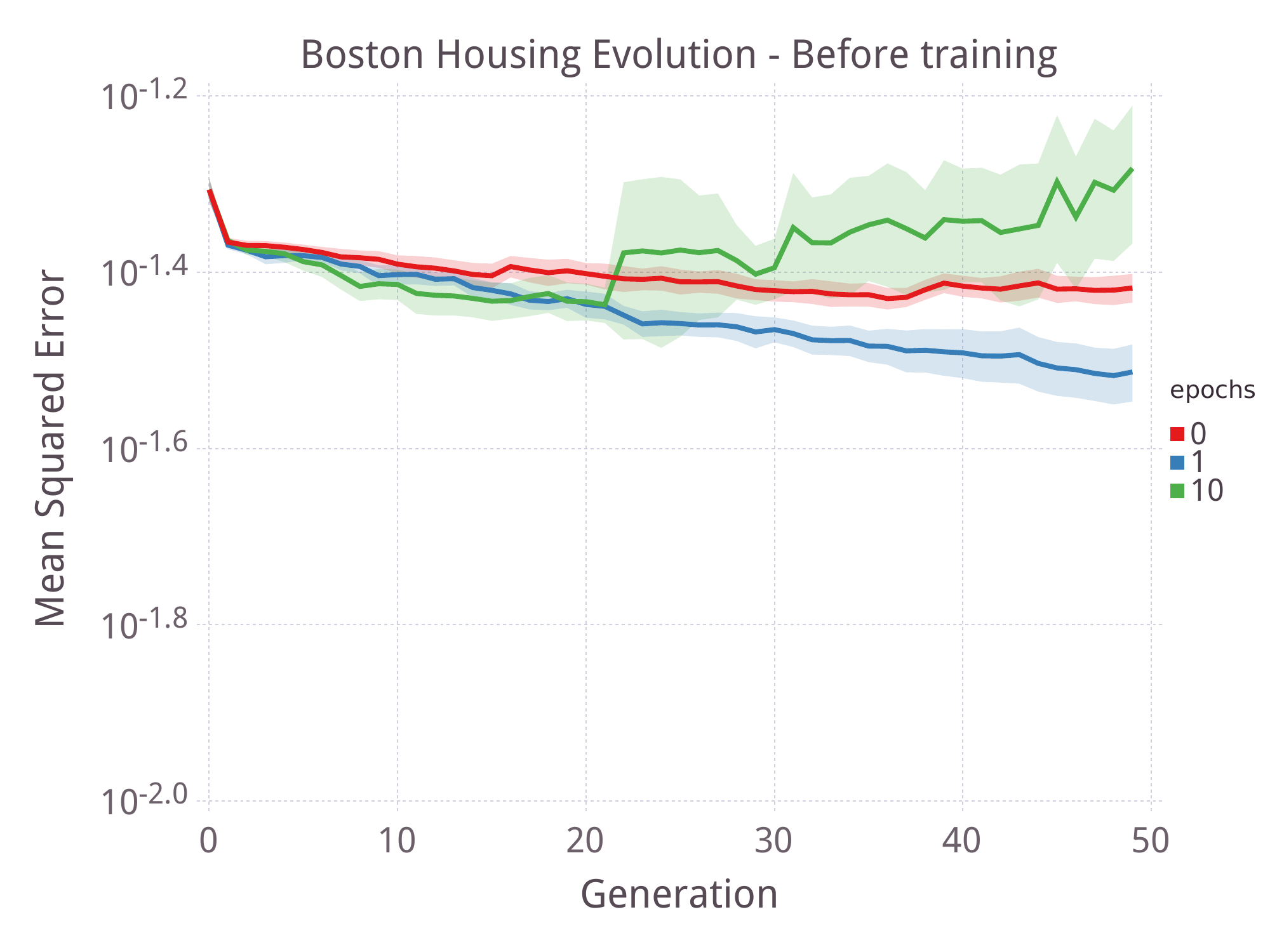}
  \includegraphics[width=0.5\textwidth]{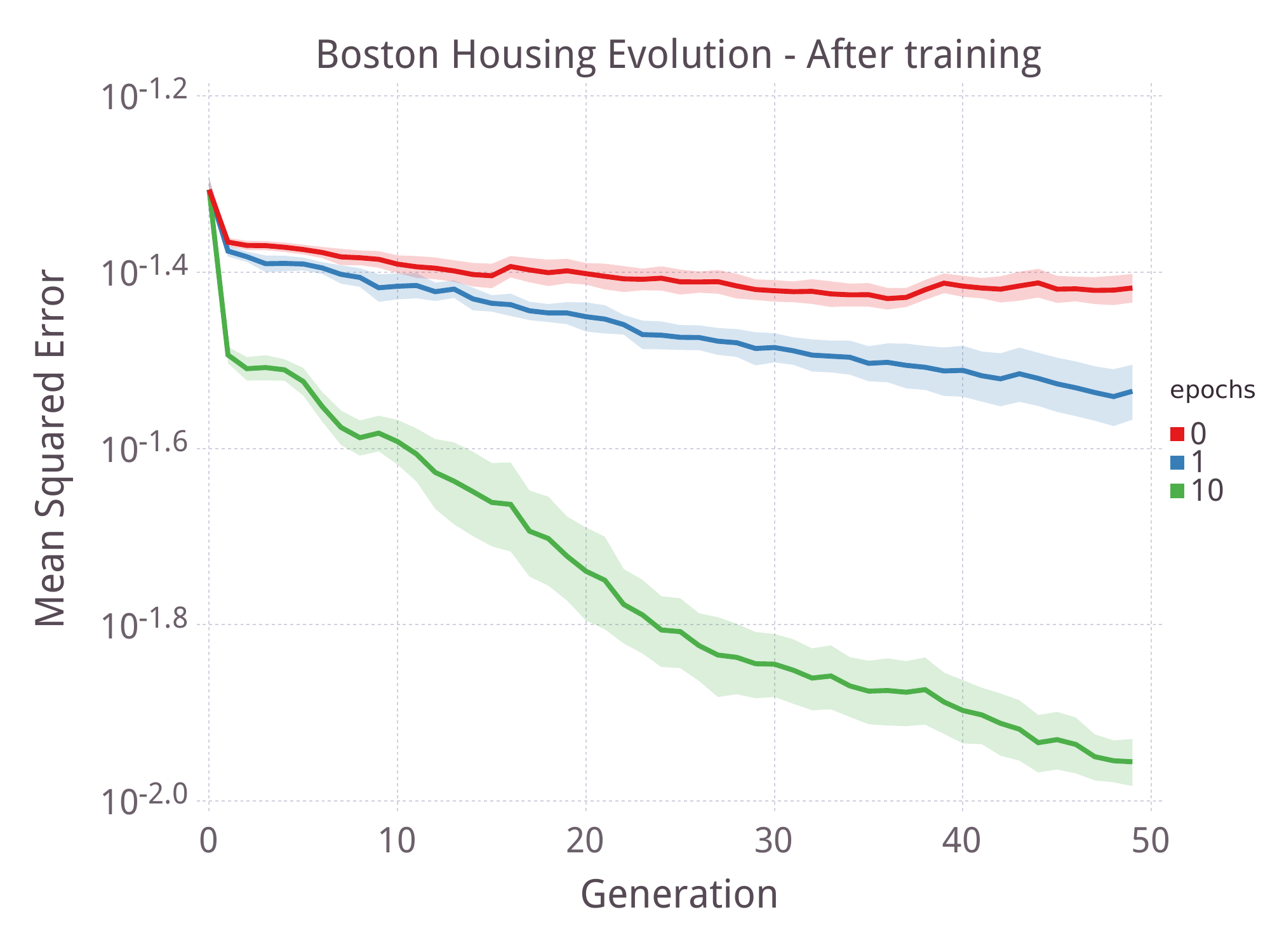}\\
  \caption{GRN evolution using different epochs for fitness. The baseline evolution is
    indicated as 0 training epochs. On the left is MSE before training; on the
    right is MSE after training. Ribbons indicate one standard deviation over 10
    trials.}
  \label{fig:evolution}
\end{figure*}

\section{Results}
\label{sec:results}

Overall, the results demonstrate that evolution can allow learning to reach more
optimal points in the search space that elude learning when performed on random
GRNs. Together, learning and evolution reach similar performance to fully
converged learners.




\autoref{fig:evolution} shows the MSE of the best individuals from each
evolution, before and after training. Evolution with 10 epochs of training shows
that using trained MSE as evolutionary fitness starts to constitute a different
evolutionary task. The error of GRNs before training in this evolution converges
above evolution-only. However,
the decreasing trained error shows that evolution is improving these GRNs.
Instead of improving initial fitness on the regression task, the learning
capability, or ``learnability'', of the population is increased over time.

Finally, we compare the trained GRNs from the 10 epoch evolution to other
models. The GRN performs better than
an RNN layer of the same size, and better than some trials of a dense neural
network. An interesting comparison is the random GRN with 50 regulatory proteins
and the dense layer, which has 60 hidden nodes. Both have a wide distribution
based on initial conditions, but are capable of very good error results. This
again demonstrates the importance of initialization for learning.

\begin{figure}[h]
  \centering
  \includegraphics[width=0.6\textwidth]{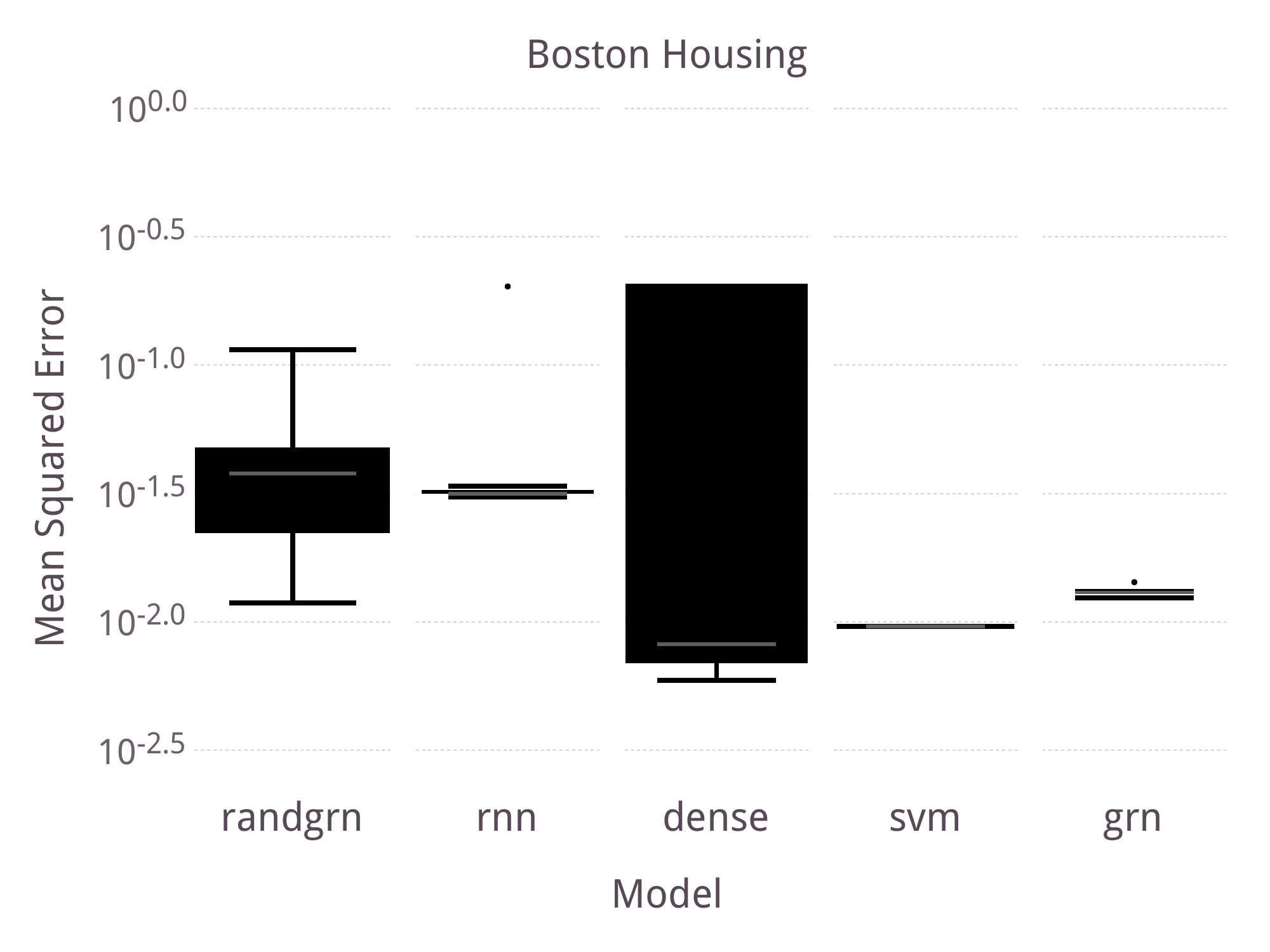}
  \caption{Test MSE of the best GRNs from the 10 epoch evolution and of other
    trained models}
  \label{img:comparison}
\end{figure}



\section{Discussion}
\label{sec:discussion}

In this work, we have demonstrated that a differentiable GRN can be used as a
layer in a learning model. While learning was only performed on individual
layers in this work, the Keras GRN layer can easily be used in conjunction with
other layer types. We believe this model is one of the first successful
implementations of an evolutionary layer in a modern deep learning framework,
and can provide an alternative to RNN layers. We are also interested in
understanding the capabilities of models with multiple GRN layers, stacked
similarly to deep neural network layers.


This work also explored the complex relationship between learning and evolution.
The Baldwin Effect is not a well-understood phenomena for artificial evolution,
and there are many open questions related to it. In this work, we chose constant
training periods for simplicity, but observed cases where learning was
unnecessary for certain individuals, or where more learning could be beneficial.
An automatic process could instead determine the length of training for each
individual, or each generation. Understanding how such a process impacts the
Baldwin Effect is intended as future work.

Also for the sake of simplicity, this work used the same evolutionary fitness
for the baseline evolution and evolution with learning. This may result in
overfitting on the training set used in evolution. Instead, one could use a
validation score as the evolutionary fitness for learning runs. This opens the
broader question of designing evolutionary goals specifically for learning
individuals. Metrics to measure the learning capacity, such as the average
increase over epoch, could be used. The results demonstrate that such a fitness
measure may be used implicitly by evolution, but a designed metric for learning
may be even more effective.

Finally, the GRN benefits from having an evolvable encoding which is also now
differentiable, with no mapping from genotype to phenotype for learning. This
means the learned weights can be directly returned to the genome in a Lamarckian
evolution scheme. Preliminary results with Lamarckian evolution demonstrate that
this is a powerful search mechanism and the use of the GRN model in Lamarckian
evolution will be the topic of future study.

\subsection*{Acknowledgments}
This work is supported by ANR-11-LABX-0040-CIMI, within programme ANR-11-IDEX-0002-02.

\bibliographystyle{apalike}
\bibliography{main} 

\end{document}